%% file: fastBP-TR.tex
\documentclass{article}
\usepackage{epsfig}
\usepackage{amsmath}
\usepackage{amssymb}
\usepackage{subfigure}
\usepackage{color}

\newcommand{\argmin}{\arg \min}

\newcommand{\deri}[2]{\frac{\partial  #1}{\partial #2}}
\newcommand{\loss}{{\cal  L}}
\newcommand{\Real}{{\cal R}}

\let\oldsubsection\subsection
\renewcommand{\subsection}[1]{\oldsubsection*{#1}}
\long\def\symbolfootnote[#1]#2{\begingroup%
\def\thefootnote{\fnsymbol{footnote}}\footnote[#1]{#2}\endgroup}
\title{Fast Inference in Sparse Coding Algorithms\\ with Applications to Object Recognition}

\date{December 4, 2008}

\author{
Koray Kavukcuoglu \hspace{0.5cm} Marc'Aurelio Ranzato \hspace{0.5cm} Yann LeCun\\
\\
Department of Computer Science\\
Courant Institute of Mathematical Sciences\\
New York University, New York, NY 10003 \\
\texttt{ \{koray,ranzato,yann\}@cs.nyu.edu} \\
}

\begin{document}

\maketitle

\begin{center}
  {\large
    Computational and Biological Learning Laboratory\\
    Technical Report\\
  }
  \medskip{}
  CBLL-TR-2008-12-01\symbolfootnote[2]{Presented at OPT 2008 Optimization for Machine Learning Workshop, Neural Information Processing Systems, 2008}

\end{center}
\bigskip{}
\begin{abstract}
  Adaptive sparse coding methods learn a possibly overcomplete set of
  basis functions, such that natural image patches can be
  reconstructed by linearly combining a small subset of these
  bases. The applicability of these methods to visual object
  recognition tasks has been limited because of the prohibitive cost
  of the optimization algorithms required to compute the sparse
  representation.  In this work we propose a simple and efficient
  algorithm to learn basis functions. After training, this model also
  provides a fast and smooth approximator to the optimal
  representation, achieving even better accuracy than exact sparse
  coding algorithms on visual object recognition tasks.
\end{abstract}

\input{introduction.tex}
\input{algorithm.tex}
\input{experiments.tex}

\section{Summary and Future Work}
Sparse coding algorithms can be used as pre-processor in many vision
applications and, in particular, to extract features in object
recognition systems. To the best of our knowledge, no sparse coding
algorithm is computationally efficient because inference involves some
sort of iterative optimization.  We showed that sparse codes can
actually be approximated by a feed-forward regressor without
compromising the recognition accuracy, but making the recognition
process very fast and suitable for use in real-time systems. We
proposed a very simple algorithm to train such a regressor.

In the future, we plan to train the model convolutionally in order to
make the sparse representation more efficient, and to build
hierarchical deep models by sequentially replicating the model on the
representation produced by the previous stage as successfully proposed
in~\cite{Hinton-DeepAutoencoder}.

\newenvironment{bibli}{
  \setlength{\parskip}{0pt}
  \setlength{\parsep}{0pt}
}

\begin{bibli}
{\small \bibliographystyle{unsrt}
  \bibliography{objrec,bib_cvpr07,bib-lecun,bib-objrec,bib-iccv07}}
\end{bibli}

\end{document}

%% file: introduction.tex
\section{Introduction}
Object recognition is one of the most challenging tasks in computer
vision. Most methods for visual recognition rely on handcrafted
features to represent images.  It has been shown that making these
representations adaptive to image data can improve performance on
vision tasks as demonstrated in~\cite{lecun-98} in a supervised
learning framework and in~\cite{elad-cvpr-06,ranzato-cvpr-07} using
unsupervised learning. In particular, learning sparse representations
can be advantageous since features are more likely to be linearly
separable in a high-dimensional space and they are more robust to
noise. Many sparse coding algorithms have been shown to learn good
local feature extractors for natural
images~\cite{olshausen-field-97,ksvd,mairal-cvpr-08,lee-nips-06,ranzato-06}. However,
application of these methods to vision problems has been limited due
to prohibitive cost of calculating sparse
representations for a given image~\cite{mairal-cvpr-08}.

In this work, we propose an algorithm named Predictive Sparse
Decomposition (PSD) that can simultaneously learn an overcomplete
linear basis set, and produce a smooth and easy-to-compute
approximator that predicts the optimal sparse representation.
Experiments demonstrate that the predictor is over 100 times faster
than the fastest sparse optimization algorithm, and yet produces
features that yield better recognition accuracy on visual object
recognition tasks than the optimal representations produced through
optimization.

\subsection{1.1 Sparse Coding Algorithms}
Finding a representation $Z \in \Real^m$ for a given signal $Y \in
\Real^n$ by linear combination of an overcomplete set of basis
vectors, columns of matrix $B \in \Real^{n \times m}$ with $m>n$, has
infinitely many solutions. In optimal sparse coding, the problem is
formulated as:
\begin{eqnarray}
  \min ||Z||_0 {\;\;\rm  s.t.\;\;} Y=BZ
\label{eq:L0}
\end{eqnarray}
where the $\ell^0$ ``norm'' is defined as the number of non-zero
elements in a given vector.  Unfortunately, the solution to this
problem requires a combinatorial search, intractable in
high-dimensional spaces. Matching Pursuit methods~\cite{Mallat:1993bs}
offer a greedy approximation to this problem. Another way to
approximate this problem is to make a convex relaxation by turning the
$\ell^0$ norm into an $\ell^1$ norm~\cite{chen99atomic}.  This
problem, dubbed Basis Pursuit in the signal processing community, has
been shown to give the same solution to eq.~(\ref{eq:L0}), provided that
the solution is sparse enough~\cite{donoho-sparse}.  Furthermore, the
problem can be written as an unconstrained optimization problem:
\begin{eqnarray}
  \loss(Y,Z;B)=\frac{1}{2} ||Y-BZ||_2^2 + \lambda||Z||_1
\label{eq:bp}
\end{eqnarray}
This particular formulation, called Basis Pursuit Denoising, can be
seen as minimizing an objective that penalizes the reconstruction
error using a linear basis set and the sparsity of the corresponding
representation. Many recent works have focused on efficiently solving
the problem in
eq.~(\ref{eq:bp})~\cite{efron02least,ksvd,lee-nips-06,Murray,ThreshCircuit,mairal-cvpr-08}.
Yet, inference requires running some sort of iterative minimization
algorithm that is always computationally expensive.

Additionally, some algorithms are also able to {\em learn} the set of
basis functions.  The learning procedure finds the $B$ matrix that
minimizes the same loss of eq.~(\ref{eq:bp}).  The columns of $B$ are
constrained to have unit norm in order to prevent trivial solutions
where the loss is minimized by scaling down the coefficients while
scaling up the bases.  Learning proceeds by alternating the
optimization over $Z$ to infer the representation for a given set of
bases $B$, and the minimization over $B$ for the given set of optimal
$Z$ found at the previous step. Loosely speaking, basis functions
learned on natural images under sparsity constraints are localized
oriented edge detectors reminiscent of Gabor wavelets.

%% file: algorithm.tex
\section{The Algorithm}

In order to make inference efficient, we train a non-linear regressor
that maps input patches $Y$ to sparse representations $Z$. We consider
the following nonlinear mapping:
\begin{eqnarray}
  F(Y;G,W,D) = G \tanh( W Y + D) \label{eq:regressor}
\end{eqnarray}
where $W \in \Real^{m \times n}$ is a filter matrix, $D \in \Real^m$
is a vector of biases, $\tanh$ is the hyperbolic tangent
non-linearity, and $G\in \Real^{m \times m}$ is a diagonal matrix of
{\em gain} coefficients allowing the outputs of $F$ to compensate for
the scaling of the input, given that the reconstruction performed by
$B$ uses bases with unit norm.  Let $P_f$ collectively denote the
parameters that are learned in this predictor, $P_f=\{G,W,D\}$.  The
goal of the algorithm is to make the prediction of the regressor,
$F(Y;P_f)$ as close as possible to the optimal set of coefficients:
$Z^* = \argmin_Z \loss(Y,Z;B)$ in eq.~(\ref{eq:bp}).  This
optimization can be carried out separately {\em after} the problem in
eq.~(\ref{eq:bp}) has been solved.  However, training becomes much
faster by {\em jointly} optimizing the $P_f$ and the set of bases $B$
all together. This is achieved by adding another term to the loss
function in eq.~(\ref{eq:bp}), enforcing the representation $Z$ to be
as close as possible to the feed-forward prediction $F(Y;P_f)$:
\begin{eqnarray}
  \loss(Y,Z;B,P_f) = \|Y - BZ \|_2^2 + \lambda \|Z\|_1 + \alpha \|Z - F(Y;P_f) \|_2^2 
\label{eq:loss}
\end{eqnarray}
Minimizing this loss with respect to $Z$ produces a representation
that simultaneously reconstructs the patch, is sparse, and is not too
different from the predicted representation. If multiple solutions to
the original loss (without the prediction term) exist, minimizing this
compound loss will drive the system towards producing basis functions
and optimal representations that are easily predictable. After
training, the function $F(Y;P_f)$ will provide good and smooth
approximations to the optimal sparse representations. Note that, a
linear mapping would not be able to produce sparse representations
using an overcomplete set because of the non-orthogonality of the
filters, therefore a non-linear mapping is required.

\subsection{2.1 Learning}
The goal of learning is to find the optimal value of the basis
functions $B$, as well as the value of the parameters in the regressor
$P_f$.  Learning proceeds by an on-line block coordinate gradient
descent algorithm, alternating the following two steps for each
training sample $Y$:
\begin{enumerate}
\item keeping the parameters $P_f$ and $B$ constant, minimize
  $\loss(Y,Z;B,P_f)$ of eq.~(\ref{eq:loss}) with respect to $Z$,
  starting from the initial value provided by the regressor
  $F(Y;P_f)$.  In our experiments we use gradient descent, but any
  other optimization method can be used;
\item using the optimal value of the coefficients $Z$ provided by the
  previous step, update the parameters $P_f$ and $B$ by one step of
  stochastic gradient descent; The update is: $U \leftarrow U - \eta
  \deri{\loss}{U}$, where $U$ collectively denotes $\{P_f,B\}$ and
  $\eta$ is the step size. The columns of $B$ are then re-scaled to
  unit norm.
\end{enumerate}
Interestingly, we recover different algorithms depending on the value
of the parameter $\alpha$:
\begin{itemize}
\item {\it $\alpha = 0$}.  The loss of eq.~(\ref{eq:loss}) reduces to
  the one in eq.~(\ref{eq:bp}). The learning algorithm becomes similar
  to Olshausen and Field's sparse coding
  algorithm~\cite{olshausen-field-97}. The regressor is trained {\em
    separately} from the set of basis functions $B$.
\item {\it $\alpha \in (0,+\infty)$}.  The parameters are updated
  taking into account also the constraint on the representation, using
  the same principle employed by SESM training~\cite{ranzato-nips07},
  for instance.
\item {\it $\alpha \rightarrow +\infty$}. The additional constraint on
  the representation (the third term in eq.~(\ref{eq:loss})) becomes
  an equality, i.e. $Z = F(Y;P_f)$, and the model becomes similar to
  an auto-encoder neural network with a sparsity regularization term
  acting on the internal representation $Z$ instead of a
  regularization acting on the parameters $P_f$ and $B$.
\end{itemize}
In this paper, we always set $\alpha = 1$. However,
sec.~\ref{sec:experiments} shows that training the regressor after
training the set of bases $B$ yields similar performance in terms of
recognition accuracy.  When the regressor is trained afterwards, the
approximate representation is usually less sparse and the overall
training time increases considerably.  Finally, additional experiments
not reported here show that training the system as an auto-encoder
($\alpha \rightarrow +\infty$) provides a very fast and efficient
algorithm that can produce good representations when the
dimensionality of the representation is not much greater than the
input dimensionality, i.e. $m \simeq n$. When the sparse
representation is highly overcomplete the block-coordinate descent
algorithm with $\alpha \in (0,+\infty)$ provides better features.

\subsection{2.2 Inference}
Once the parameters are learned, inferring the representation $Z$ can
be done in two ways. \\
{\bf Optimal inference} consists of setting the representation to $Z^*
= \argmin_z \loss$, where $\loss$ is defined in eq.~(\ref{eq:loss}),
by running an iterative gradient descent algorithm involving two
possibly large matrix-vector multiplications at each iteration (one
for computing the value of the objective, and
one for computing the derivatives through $B$). \\
{\bf Approximate inference}, on the other hand sets the representation
to the value produced by $F(Y;P_f)$ as given in
eq.~(\ref{eq:regressor}), involving only a forward propagation through
the regressor, i.e. a single matrix-vector multiplication.

%% file: experiments.tex
\section{Experiments} \label{sec:experiments}

First, we demonstrate that the proposed algorithm (PSD) is able to
produce good features for recognition by comparing to other
unsupervised feature extraction algorithms, Principal Components
Analysis (PCA), Restricted Boltzman Machine
(RBM)~\cite{Hinton-rbm-cd}, and Sparse Encoding Symmetric Machine
(SESM)~\cite{ranzato-nips07}. Then, we compare the recognition
accuracy and inference time of PSD feed-forward approximation to
feature sign algorithm~\cite{lee-nips-06}, on the Caltech 101
dataset~\cite{feifei_cvpr04}. Finally we investigate the stability of
representations under naturally changing inputs.

\subsection{3.1 Comparison against PCA, RBM and SESM on the MNIST}
\begin{figure}[tbh]
  \begin{centering}
    \includegraphics[width=1\textwidth,height=0.25\textwidth]{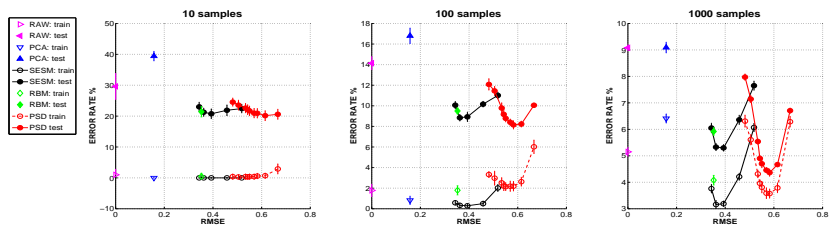}
    \par\end{centering}
  \caption{Classification error on MNIST as a function of
    reconstruction error using raw pixel values and, PCA, RBM, SESM
    and PSD features.  Left-to-Right : 10-100-1000 samples per class
    are used for training a linear classifier on the features.  The
    unsupervised algorithms were trained on the first 20,000 training
    samples of the MNIST dataset~\cite{MNIST}. }
  \label{fig:sparsity-error-comparison-nips07}
\end{figure}

The MNIST dataset has a training set with 60,000 handwritten digits of
size 28x28 pixels, and a test set with 10,000 digits. Each image is
preprocessed by normalizing the pixel values so that their standard
deviation is equal to 1. In this experiment the sparse representation
has 256 units. This internal representation is used as a global
feature vector and fed to a linear regularized logistic regression
classifier. Fig.~\ref{fig:sparsity-error-comparison-nips07} shows the
comparison between PSD (using feed-forward approximate codes) and,
PCA, SESM \cite{ranzato-nips07}, and RBM
\cite{Hinton-DeepAutoencoder}.  Even though PSD provides the {\bf
  worst reconstruction error}, it can achieve the {\bf best
  recognition accuracy} on the test set under different number of
training samples per class.

\subsection{3.2 Comparison with Exact Algorithms}
In order to quantify how well our jointly trained predictor given in
eq.~(\ref{eq:regressor}) approximates the optimal representations
obtained by minimizing the loss in eq.~(\ref{eq:loss}) and the optimal
representations that are produced by an exact algorithm minimizing
eq.~(\ref{eq:bp}) such as feature sign~\cite{lee-nips-06} (FS), we
measure the average signal to noise ratio\footnote{$SNR = 10 log_{10}
  (\sigma^2_{signal} / \sigma^2_{noise})$} (SNR) over a test dataset
of 20,000 natural image patches of size 9x9.  The data set of images
was constructed by randomly picking 9x9 patches from the images of the
Berkeley dataset converted to gray-scale values, and these patches
were normalized to have zero mean and unit standard deviation.  The
algorithms were trained to learn sparse codes with 64
units\footnote{Principal Component Analysis shows that the effective
  dimensionality of 9x9 natural image patches is about 47 since the
  first 47 principal components capture the 95\% of the variance in
  the data. Hence, a 64-dimensional feature vector is actually an
  overcomplete representation for these 9x9 image patches.}.

We compare representations obtained by ``PSD Predictor'' using the
{\em approximate} inference, ``PSD Optimal'' using the {\em
  optimal} inference, ``FS'' minimizing eq.~(\ref{eq:bp}) with~\cite{lee-nips-06}, 
and ``Regressor'' that is separately trained to approximate the exact optimal codes produced by
FS.  The results given in table~\ref{table:snr} show that PSD direct
predictor achieves about the same SNR on the true optimal
sparse representations produced by FS, as the Regressor
that was trained to predict these representations.
\begin{table}[bt]
  \caption{Comparison between representations produced by FS~\cite{lee-nips-06} and PSD.
 In order to compute the SNR, the noise is defined as $(Signal-Approximation)$.
  }
\begin{centering}
\begin{tabular}{|l|c|}
  \hline 
  Comparison (Signal / Approximation) & Signal to Noise Ratio (SNR)\tabularnewline
  \hline
  \hline 
  1. PSD Optimal / PSD Predictor & 8.6\tabularnewline
  \hline 
  2. FS / PSD Optimal & 5.2\tabularnewline
  \hline 
  3. FS / PSD Predictor & 3.1\tabularnewline
  \hline 
  4. FS / Regressor & 3.2\tabularnewline
  \hline
\end{tabular}
\par\end{centering}
\label{table:snr}
\end{table}

\begin{figure}[tb]
  \begin{centering}
      \includegraphics[width=0.73\textwidth]{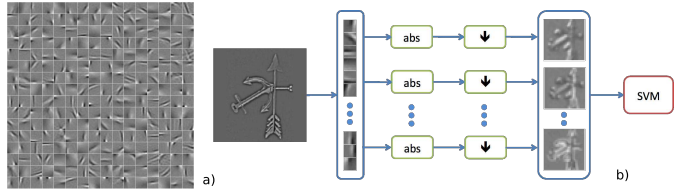}
    \par\end{centering}
  \caption{{\bf a)} 256 basis functions of size 12x12 learned by PSD,
    trained on the Berkeley dataset. Each 12x12 block is a column of
    matrix $B$ in eq.~(\ref{eq:loss}), i.e. a basis function.  {\bf
      b)} Object recognition architecture: linear adaptive filter
    bank, followed by $abs$ rectification, average down-sampling and
    linear SVM classifier.}
      \label{fig:recog_sys}
\end{figure}

Despite the lack of absolute precision in predicting the exact optimal
sparse codes, PSD predictor achieves even better performance in
recognition. The Caltech 101 dataset is pre-processed in the following
way: {\bf 1)} each image is converted to gray-scale, {\bf 2)} it is
down-sampled so that the longest side is 151 pixels, {\bf 3)} the mean
is subtracted and each pixel is divided by the image standard
deviation, {\bf 4)} the image is locally normalized by subtracting the
weighted local mean from each pixel and dividing it by the weighted
norm if this is larger than 1 with weights forming a 9x9 Gaussian
window centered on each pixel, and {\bf 5)} the image is 0-padded to
143x143 pixels.  64 feature detectors (either produced by FS or PSD
predictor) were plugged into an image classification system that {\bf A)}
used the sparse coding algorithms convolutionally to produce 64
feature maps of size 128x128 for each pre-processed image, {\bf B)}
applied an absolute value rectification, {\bf C)} computed an average
down-sampling to a spatial resolution of 30x30 and {\bf D)} used a
linear SVM classifier to recognize the object in the image (see
fig.~\ref{fig:recog_sys}(b)).  Using this system with 30 training images
per class we can achieve $53\%$ accuracy on Caltech 101 dataset.

\begin{figure}[tb]
  \begin{centering}
    \includegraphics[width=0.75\textwidth]{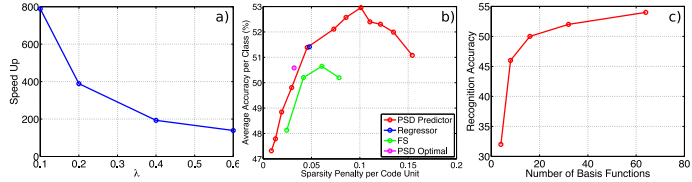}
    \par\end{centering}
 \caption{{\bf a)} Speed up for inferring the sparse representation
   achieved by PSD predictor over FS for a code with 64 units. The
   feed-forward extraction is more than 100 times faster. {\bf b)}
   Recognition accuracy versus measured sparsity (average $\ell^1$
   norm of the representation) of PSD predictor compared to the to the
   representation of FS algorithm.  A difference within 1\% is not
   statistically significant. {\bf c)} Recognition accuracy as a
   function of number of basis functions.}
 \label{fig:acc_nfilters}
\end{figure}

Since FS finds exact sparse codes, its representations are generally
sparser than those found by PSD predictor trained with the same value
of sparsity penalty $\lambda$.  Hence, we compare the recognition
accuracy against the {\em measured} sparsity level of the
representation as shown in fig.~\ref{fig:acc_nfilters}(b).  PSD is not
only able to achieve better accuracy than exact sparse coding
algorithms, but also, it does it much more efficiently.
Fig.~\ref{fig:acc_nfilters}(a) demonstrates that our feed-forward
predictor extracts features more than 100 times faster than feature
sign. In fact, the speed up is over 800 when the sparsity is set to
the value that gives the highest accuracy shown in
fig.~\ref{fig:acc_nfilters}(b).

Finally, we observe that these sparse coding algorithms are somewhat
inefficient when applied convolutionally.  Many feature detectors are
the translated versions of each other as shown in
fig.~\ref{fig:recog_sys}(a).  Hence, the resulting feature maps are
highly redundant. This might explain why the recognition accuracy
tends to saturate when the number of filters is increased as shown in
fig.~\ref{fig:acc_nfilters}(c).

\subsection{3.3 Stability}
In order to quantify the stability of PSD and FS, we investigate their
behavior under naturally changing input signals. For this purpose, we
train a basis set with 128 elements, each of size 9x9, using the PSD
algorithm on the Berkeley~\cite{Berkeley} dataset. This basis set is
then used with FS on the standard ``foreman'' test video together with
the PSD Predictor. We extract 784 uniformly distributed patches from
each frame with a total of 400 frames.  

\begin{figure}[tbh]
  \begin{centering}
    \includegraphics[width=0.71\textwidth,height=0.18\textwidth]{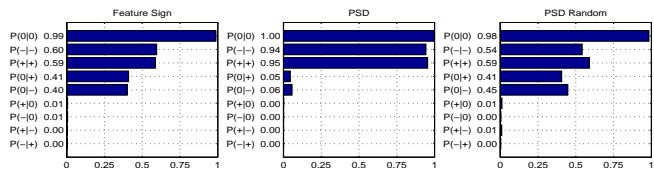}
    \par\end{centering}
  \caption{Conditional probabilities for sign transitions between two
    consecutive frames. For instance, $P(-|+)$ shows the conditional probability of a unit
    being negative given that it was positive in the previous frame.
    The figure on the right is used as baseline, showing the
    conditional probabilities computed on pairs of {\em random}
    frames.}
  \label{fig:stability}
\end{figure}
For each patch, a 128 dimensional representation is calculated using
both FS and the PSD predictor.  The stability is measured by the
number of times a unit of the representation changes its sign, either
negative, zero or positive, between two consecutive frames. Since the
PSD predictor does not generate exact zero values, we threhsold its
output units in such a way that the average number of zero units
equals the one produced by FS (roughly, only the $4\%$ of the units
are non-zero). The transition probabilities are given in
Figure~\ref{fig:stability}.  It can be seen from this figure that the
PSD predictor generates a more stable representation of slowly varying
natural frames compared to the representation produced by the exact
optimization algorithm.
